\documentclass[journal,twoside,web]{ieeecolor}
\usepackage{tmi}
\usepackage{cite}
\usepackage{amsmath,amssymb,amsfonts}
\usepackage{algorithmic}
\usepackage{graphicx}
\usepackage{textcomp}
\usepackage{xcolor,soul,framed} 
\usepackage{grffile}

\colorlet{shadecolor}{yellow}

\usepackage{mdwmath}
\usepackage{mdwtab}
\usepackage{eqparbox}
\usepackage{url}
\usepackage{verbatim}
\usepackage{threeparttable}
\newcommand{\tabincell}[2]{\begin{tabular}{@{}#1@{}}#2\end{tabular}}
\usepackage{soul}
\usepackage{multirow}

\def\BibTeX{{\rm B\kern-.05em{\sc i\kern-.025em b}\kern-.08em
    T\kern-.1667em\lower.7ex\hbox{E}\kern-.125emX}}
\begin{document}
\title{Structure-Aware Long Short-Term Memory Network for 3D Cephalometric Landmark Detection}
  \author{Runnan~Chen, Yuexin~Ma, Nenglun~Chen, Lingjie~Liu, \\
        Zhiming~Cui, Yanhong~Lin, Wenping~Wang,~\IEEEmembership{Fellow,~IEEE}
  \thanks{This work is funded by the Innovative Technology Fund of the Innovation and Technology Bureau, Hong Kong SAR.}
  \thanks{Prof. Wenping Wang, Mr. Runnan Chen, Mr. Nenglun Chen, Mr. Zhiming Cui and Miss. Yanhong Lin are with the department of computer science, The University of Hong Kong, Hong Kong SAR (e-mail: wenping@cs.hku.hk, rnchen2@cs.hku.hk, chennenglun@gmail.com, cuizm.neu.edu@gmail.com, yhlinedith@gmail.com)}
  \thanks{Mrs. Yuexin Ma, is with the school of information science and technology, ShanghaiTech University, Shanghai Engineering Research Center of Intelligent Vision and Imaging, Shanghai (e-mail: mayuexin@shanghaitech.edu.cn).}
  \thanks{Miss. Lingjie Liu, is with Max Planck Institute for Informatics, Saarbrücken (e-mail: lliu@mpi-inf.mpg.de).}
  }

\definecolor{crncolor}{rgb}{.0,0.5,0.8}
\newcommand{\CRN}[1]{{\textcolor{crncolor}{[\textbf{CRN:} #1]}}}

\maketitle

\begin{abstract}
Detecting 3D landmarks on cone-beam computed tomography (CBCT) is crucial to assessing and quantifying the anatomical abnormalities in 3D cephalometric analysis. However, the current methods are time-consuming and suffer from large biases in landmark localization, leading to unreliable diagnosis results. In this work, we propose a novel Structure-Aware Long Short-Term Memory framework (SA-LSTM) for efficient and accurate 3D landmark detection. To reduce the computational burden, SA-LSTM is designed in two stages. It first locates the coarse landmarks via heatmap regression on a down-sampled CBCT volume and then progressively refines landmarks by attentive offset regression using multi-resolution cropped patches. To boost accuracy, SA-LSTM captures global-local dependence among the cropping patches via self-attention. Specifically, a novel graph attention module implicitly encodes the landmark's global structure to rationalize the predicted position. Moreover, a novel attention-gated module recursively filters irrelevant local features and maintains high-confident local predictions for aggregating the final result. Experiments conducted on an in-house dataset and a public dataset show that our method outperforms state-of-the-art methods, achieving 1.64 mm and 2.37 mm average errors, respectively. Furthermore, our method is very efficient, taking only 0.5 seconds for inferring the whole CBCT volume of resolution 768$\times$768$\times$576.


\end{abstract}

\section{Introduction}
Cephalometric analysis is widely adopted in orthodontic and maxillofacial surgeries for diagnosis, treatment planning and evaluation\cite{gateno2011new,bettega2000simulator,hurst2007surgical}. Accurate and reproducible anatomical landmark detection is a crucial step to assess and quantify the anatomical abnormalities \cite{Chen2019cephalometric,Payer2019Integrating,chen2022semi}. Conventionally, cephalometric analysis is performed on 2D cephalograms. However, 2D cephalograms only provide a projected 2D view of real three-dimensional anatomical structures, resulting in problems such as magnification, distortion and imperfect overlaps of anatomical structures.  Recently, cone-beam computed tomography (CBCT) with three-dimensional (3D) representation of craniofacial structures overcomes the limitations of 2D cephalogram and performs more accurate cephalometric measurements, has become popular in orthodontic and maxillofacial surgeries \cite{chien2009comparison,moshiri2007accuracy,scarfe2006clinical,rossini20113d,pittayapat2014three}. However, manually landmarking in 3D CBCT is tedious, time-consuming and lacks reproducibility. Therefore, a fast, accurate, and automatic 3D landmark detection system is meaningful for clinical usages.

Efficient and accurate landmark detection on a high-resolution 3D CBCT volume is challenging due to the curse of dimensionality. Processing 3D volume data is computationally expensive as it needs more network parameters and computational resources than processing 2D images. Moreover, the prediction error is increased accordingly because the 3D volume has a much larger search space than its 2D projection. More importantly, considering the spatial relationship among landmarks is beneficial for more accurate abnormality diagnosis in cephalometric analysis. It is because that the pathology types are measured based on the spatial relationship among multiple landmarks. Therefore, the individual landmark will significantly affect the diagnosis accuracy \cite{wang2016benchmark}, i.e., if some predicted landmarks significantly deviate from the ground truth, the related pathology measurements (e.g., angles and lengths) are unreliable, resulting in an unacceptable diagnosis in cephalometric analysis.

Some traditional regression-based methods \cite{zhang2016detecting,han2014robust,han2015robust,zhang2015automatic,zhang2017joint,gao2015collaborative} are proposed in solving 3D CBCT landmark detection. They typically model the landmark appearance by handcraft features and aggregate the final prediction via regression voting. Although some achieve promising results, they are sensitive to landmarks with ambiguous appearance and artifacts, and they may fail when encountering geometrically complex landmarks. The deep learning-based methods automatically learn deep features from data and have more powerful generalization ability. Payer et al. \cite{Payer2019Integrating} integrate the spatial configurations into heatmaps for landmark regression. Zhang et al. \cite{zhang2017detecting} employ a cascaded CNN framework to regress landmarks from limited data. However, these methods are unsuitable for inputting high-resolution data (768$\times$768$\times$576) due to the huge consumption of computational resources. Lang et al. \cite{lang2020automatic} propose a multi-stage 3D heatmap regression method and utilize a graph convolution network to determine the landmark's existence. However, it is time-consuming minutes to process a CBCT volume. Besides, the lack of consideration of global structure constraint leads to large-biased landmark localization and unacceptable diagnosis results.

\begin{figure*}
  \vspace*{2ex}
  \centering
  \includegraphics[width=1\textwidth]{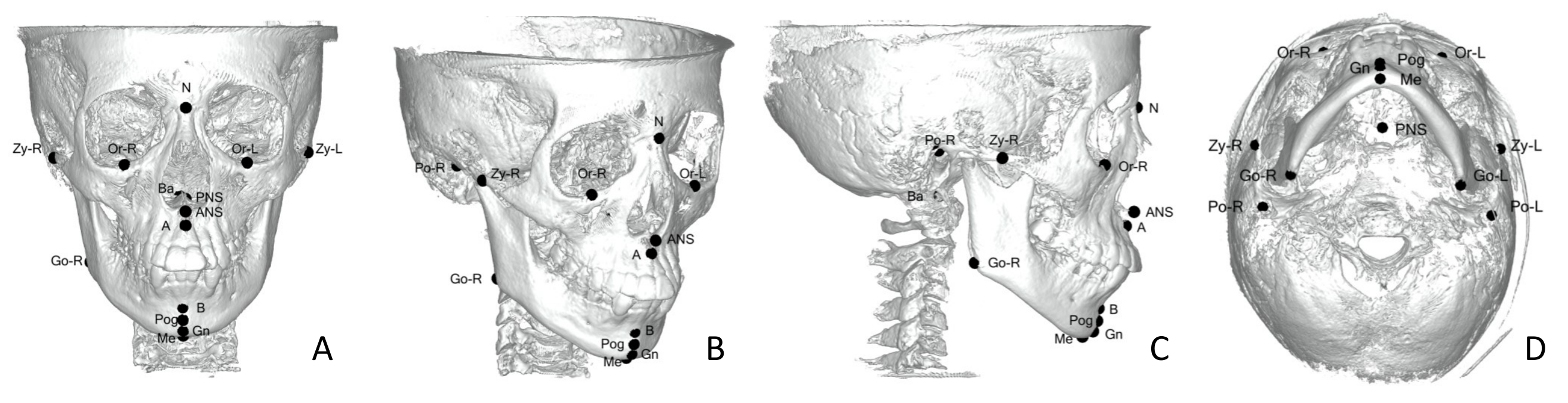}
  \caption{Seventeen anatomical landmarks on bones. They are shown in four different views: A, coronal; B, isometric; C, sagittal; D, axial. Orbitale (Or\_R, Or\_L): the most inferior point on infraorbital rim, right and left; Zygomatic (Zy\_R, Zy\_L): the most lateral point on the left and right outline of zygomatic arch; Gonion (Go\_R, Go\_L): a point in the middle of the curvature at the left and right angles of the mandible; Porion (Po\_R, Po\_L): the most superior point of anatomic external auditory meatus, right and left; Nasion (N): midsagittal point at junction of frontal and nasal bones at nasofrontal suture; Anterior nasal spine (ANS): the most anterior limit of floor of nose at tip of ANS; Posterior nasal spine (PNS): point along palate immediately inferior to pterygomaxillary fossa; Basion (Ba): the most inferior point on anterior margin of foramen magnum at base of clivus; A-point (A): the most concave point of anterior maxilla; B-point (B): the most concave point on mandibular symphysis; Pogonion (Pog): the most anterior point along curvature of chin; Menton (Me): the most inferior point along curvature of chin; Gnathion (Gn): located perpendicular on madibular symphysis midway between Pog and Me.}
  \label{fig:data_sample}
  \vspace{-1ex}
\end{figure*}

In this paper, we propose a novel framework that captures global-local dependence among landmarks for efficient and accurate 3D anatomical landmark detection on CBCT. Processing a high-resolution CBCT volume with a neural network is very computationally expensive, and the space consumption would exceed the commonly used GPU memory (11G in GTX 1080 Ti). To solve this issue, we design a framework in two stages to keep the whole processing efficient and effective. In the first stage, we adopt heatmap regression to locate coarse landmarks on the down-sampled data, which references the following refinement. In the second stage, a novel structure-aware LSTM progressively refines the landmarks via attentive offset regression using the multi-resolution cropped patches. In each iteration, SA-LSTM predicts the offsets between the patch centres and ground truth landmarks. Thus, the predicted landmarks are the sum of the predicted offsets and the patch centres, which are regarded as the central points to crop patches for the next prediction. 

Besides, the consistency of the spatial relationship among anatomical landmarks is the inherent property in the skull CBCT, e.g., the related position between landmark pairs is consistent in humans. To this end, we exploit the global-local dependence among cropped patches to boost accuracy. On the one hand, a novel graph attention module (GAM) produces structure-aware embedding to implicitly encode the global structure information to ensure predicted landmark position rational. On the other hand, the self-attention-gated module (SAG) filters irrelevant local features and maintains the high-confident local predictions for aggregating the final result. In this way, the cooperation of global structure regulation by GAM and the local prediction aggregation by SAG leads to accurate landmark detection in CBCT.

Extensive experiments conducted on an in-house CBCT dataset and a public CT dataset verify the effectiveness of our method. It achieves 1.64 mm average errors in the in-house dataset with 89 CBCT volumes, and 74.28\% of landmarks are within 2 mm of clinically acceptable errors. Moreover, by applying SA-LSTM, our method outperforms the state of the art methods. The improvements are 7\%, 2\% and 1\% in the successful detection rates of 2 mm, 2.5 mm, and 3 mm, respectively. Our method also outperforms other deep learning-based methods and is comparable with a traditional regression-based method on a public CT dataset with 33 samples. The average error is 2.37 mm, and 56.36\% of landmarks are within the 2 mm errors. Furthermore, our method is very efficient. The inference time for processing a CBCT volume is 0.5 seconds on a GTX 1080 Ti GPU or 8 seconds on a Xeon(R) Silver 4108 CPU.

The contributions of our work are as follows.
\begin{itemize}
\item {We develop a novel framework that captures global-local dependence among landmarks for efficient and accurate 3D anatomical landmark detection on CBCT.}
\item {We propose a novel self-attention-gated module for aggregating fine-scale landmark predictions.}
\item {We propose a novel graph attention module to encode global structure information for implicit shape regulation.}
\item {Our method outperforms the state-of-the-art methods on an in-house skull CBCT dataset and a public CT dataset.}
\end{itemize}

\section{RELATED WORK}
3D Anatomical landmarks in the cephalometric analysis have clinical definitions (Figure \ref{fig:data_sample}), which reflects the patient’s dental, skeletal, and facial relationship. In this paper, we focus on 17 most commonly used landmarks on the bone \cite{montufar2018hybrid,gupta2016accuracy}. There are elaborative efforts towards solving 3D anatomical landmark detection, including knowledge-based, model-based, and deep learning-based methods.
\subsection{Knowledge-based method}

The knowledge-based methods \cite{gupta2016accuracy,shahidi2013accuracy,makram2014reeb} utilize domain knowledge to model a geometrical description for detecting particular landmarks. However, they yield ambiguous results when encountering geometrically complex landmarks. 
\subsection{Model-based method}
The model-based approaches first build a statistical model of the landmark geometric feature and then apply a template-matching algorithm to search the best-fit region. Montufar et al. \cite{montufar2018hybrid} present a hybrid method that combines active shape models and a 3D knowledge-based searching algorithm. However, the mathematical formulae are sensitive to individual variations when applying to a complex 3D craniofacial model.

\subsection{Regression-based method}
Regression-based landmark method \cite{zhang2016detecting,han2014robust,han2015robust,zhang2015automatic,zhang2017joint,gao2015collaborative} exploits context information to localize landmark. In the training stage, a regression forest is learned to predict the offset of arbitrary points to the ground truth landmark according to the surrounding appearance. In the testing stage, they iteratively predicate the offsets from the current point to ground truth \cite{han2015robust} or ensemble the multiple predictions via collaborative regression-voting \cite{gao2015collaborative}. However, they typically model the appearance via handcraft features such as 3D Haar-like features, which are sensitive to landmarks with ambiguous appearance and artifacts. They also may fail when encountering geometrically complex landmarks.

\subsection{Deep Learning-based method}
As the deep learning technique \cite{lecun2015deep} achieves remarkable progress in medical image analysis \cite{litjens2017survey}, it is applied in the 3D anatomical landmark detection field and has achieved promising results. Yang et al. \cite{yang2015automated} propose a slice-based 2.5D approach to detect anatomical landmarks in 3D MRI scans. Zheng et al. \cite{zheng20153d} adopt a patch-based method that uses two cascaded multilayer perceptrons (MLPs) to detect landmarks. Ghesu et al. \cite{ghesu2017multi} explore a deep reinforcement learning-based method for landmark localization. However, the spacial relationship of landmarks (e.g. the symmetry, angle, distance), which is important for cephalometric analysis, is not considered in these methods. Payer et al. \cite{Payer2019Integrating} integrate the spatial configurations into heatmaps for landmark regression. Zhang et al. \cite{zhang2017detecting} employ a cascaded CNN framework, i.e., a CNN is used for 3D landmark regression, followed by another CNN, which correlates detected landmarks. However, these methods are unsuitable for large-size input data because of the huge consumption of computational resources and will lose accuracy if performing on down-sampled input data. Some methods detect landmarks based on the segmented 3D skull \cite{lee2019automatic} or segmented 3D mandible \cite{zhang2015automatic,torosdagli2018deep,lian2020multi}. However, segmentation of 3D skull and 3D mandible on CBCT volumes are also not easy, which leads to propagation errors for 3D landmarking. Land et al. \cite{lang2020automatic} utilize a two-stage approach to localize the landmarks with a graph convolution network to determine the existence of each landmark. However, they utilize 3D mask R-CNN to detect landmarks which is very computational expensive that takes minutes to infer a CBCT volume. Besides, the lacking consideration of the global structure constraint leads to large-biased landmark localization. Our SA-LSTM captures global-local dependence to progressively refine landmark via attentive offset regression, overcoming the limitations above and faster and more accurate than the current methods.

\begin{figure*}
  \centerline{\includegraphics[width=1\textwidth]{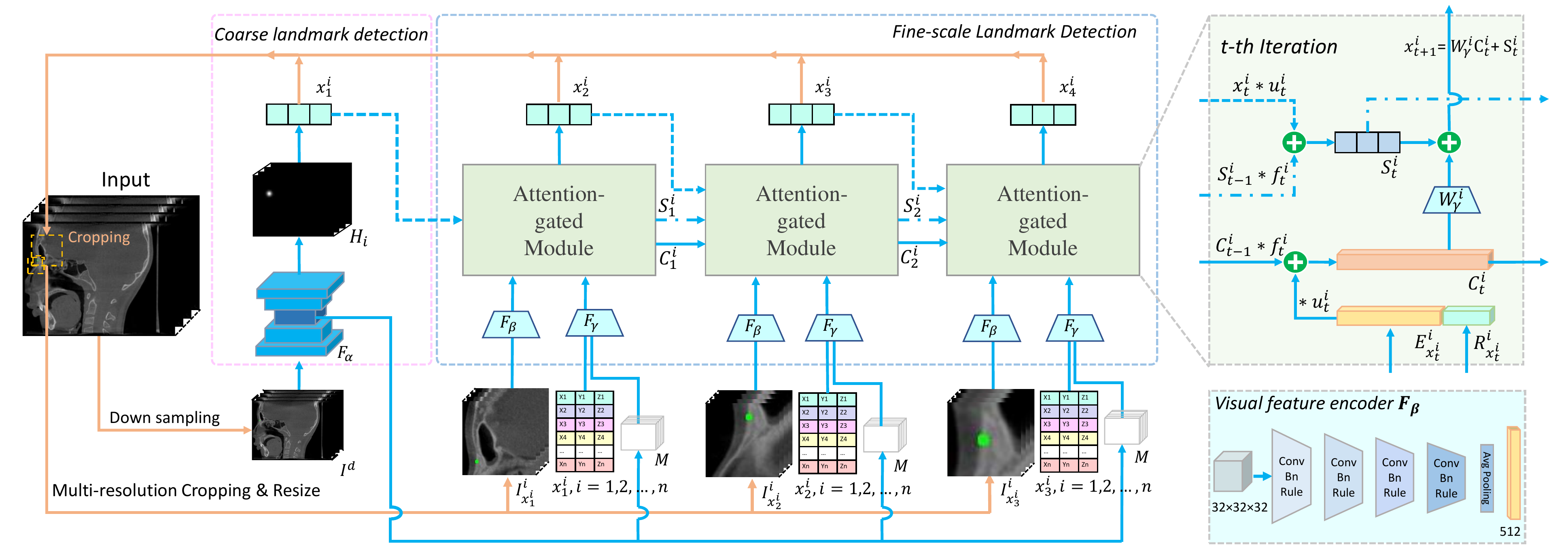}}
  \caption{Illustration of our framework for the $i$-th landmark prediction. Our framework progressively performs landmark detection. The coarse landmarks are predicted on down-sampled data, and the fine-scale landmarks are recursively predicted on the patches cropped from the original CBCT volume by the SA-LSTM. We take the last iteration prediction $x^{i}_{4}$ as the final result. $\mathcal{F}{\alpha}$ is used for coarse landmark detection (section III.A). $\mathcal{F}_{\beta}$ is the visual feature encoder, and $\mathcal{F}_{\gamma}$ is the graph attention module to produce the structure-aware embedding of a 3D patch. $\mathcal{W}^{i}_{\gamma}$ predicts the offset between the patch central and the ground truth for the $i$-th landmark (section III.B). Note that $\mathcal{F}{\alpha}$,  $\mathcal{F}_{\beta}$ and $\mathcal{F}_{\gamma}$ are shared for all landmark predictions.}
  \label{fig:framework}
  \vspace{-2ex}
\end{figure*}

\section{Method}

Given a CBCT scan $I$, our method aims at automatically detecting $n$ anatomical landmarks. The framework is illustrated in Fig. \ref{fig:framework}. In the first stage, the coarse landmarks are predicted on a down-sampled CBCT volume via heatmap regression. In the second stage, fine-scale landmarks are predicted based on the cropped multi-resolution 3D patches around the coarse landmarks. In the following, we present details of these two stages and give more insights into our method.
\subsection{Coarse landmark detection}
The original size of CBCT volume is too huge to be processed by even a primary neural network. For example, it will probably exceed the limitation of GPU memory if a volume with the size of 768 $\times$ 768 $\times$ 576 processed by a U-Net. Therefore, a volume is down-sampled by $k$ times and is token as the network input. It is worth noting that some landmarks on tiny structure bone may become indistinguishable or even missing in the down-sampled volume data, e.g., the landmark of PNS, Ba (Figure 1). However, as the human skull is highly structured, the contextual information central to the landmark remains in the down-sampled volume, guiding the network to roughly estimate the landmark position. Therefore, the network exploits global contextual information to detect coarse landmarks, narrowing the fine-scale search space in the second stage. 

The network structure is shown in Fig. \ref{fig:coarse_GAM}, where the input of the network $\mathcal{F}_{\alpha}$ is a down-sampled volume, and the output is $n$ 3D heatmaps. Each heatmap represents a landmark probability distribution, indicating the landmark's rough position. The $i$-th coarse landmark $x^{i}_{1}\in\mathcal{R}^3$ is calculated by the integral operation on the $i$-th heatmap $\mathcal{H}_{i}$.
\begin{equation}\label{equ:integralFunction}
x^{i}_{1} = \frac{1}{\gamma_i}\sum_{\rho\in \mathcal{H}_{i}}\mathcal{H}_{i}(\rho)*\rho, \gamma_i = \sum_{\rho\in \mathcal{H}_{i}}\mathcal{H}_{i}(\rho),
\end{equation}
where $\mathcal{H}_{i}(\rho)$ is the value of the $i$-th heatmap $\mathcal{H}_{i}$ at the voxel $\rho\in\mathcal{R}^3$, and $\gamma_i$ is the spacial normalization term.

\subsection{Fine-scale Landmark Detection}
The down-sampled volume loses landmark geometrical details, which results in low detection accuracy in the first stage. To overcome the limitation, we crop the 3D patches nearby the coarse landmarks from the original volume without losing geometry details for fine-scale landmark detection. Regression-voting \cite{lindner2014robust} is a common strategy that randomly generates mass patches and aggregates multiple predictions. However, processing mass 3D patches is time-consuming and costs huge computational resources, especially on 3D volume. Moreover, due to the lack of global-local dependence modelling, some predicted landmarks may have a great deviation from the expected position, which results in an unacceptable diagnosis in cephalometric analysis. Therefore, we propose SA-LSTM, a novel Structure-Aware Long Short-Term Memory Network with attention gates, to solve the above issues. Firstly, it efficiently produces a sequential prediction progressively. Secondly, a graph attention module encodes the global shape structure to make the prediction more reasonable. Lastly, the attention-gated module captures local dependences among patches for aggregating the final prediction. We describe how SA-LSTM performs fine-scale landmark detection in the following.

\subsubsection{\textbf{Formulation}}
As shown in Figure \ref{fig:framework}, SA-LSTM recursively produces sequential predictions $x^{i}_{t+1}(t=1,2,...,T)$ for the $i$-th landmark, where $T$ is the number of iteration. In the $t$-th iteration, the network predicts landmarks $x^{i}_{t+1}(i=1,2,...,n)$ by inputting current patches $I^{i}_{x^{i}_{t}}(i=1,2,...,n)$, where $I^{i}_{x^{i}_{t}}$ are centered on $x^{i}_{t}$. Then the predicted landmarks $x^{i}_{t+1}(i=1,2,...,n)$ are regarded as the centrals for cropping other patches $I^{i}_{x^{i}_{t+1}}(i=1,2,...,n)$ for the next iterative prediction. Note that the 3D patch is represented as an embedding vector, which greatly reduces the computational burden. Moreover, we adopt multi-resolution cropping strategy for data augmentation, i.e., $I^{i}_{x^{i}_{1}}\geq I^{i}_{x^{i}_{2}}\geq I^{i}_{x^{i}_{3}}...$ in terms of the resolution. In the following, We take the prediction of the $i$-th landmark as an example to demonstrate the formulation.
\begin{align} 
    &C^{i}_{t} = f^{i}_{t}*C^{i}_{t-1} + u^{i}_{t}*(E^{i}_{x^{i}_{t}} \copyright R^{i}_{x^{i}_{t}}),& \label{equ:featureState} \\
    &S^{i}_{t} = f^{i}_{t}*S^{i}_{t-1} + u^{i}_{t}*x^{i}_{t},& \label{equ:spatialState} \\
    &x^{i}_{t+1} = \mathcal{W}^{i}_{\gamma}C^{i}_{t} + S^{i}_{t},& \label{equ:prediction}
\end{align}
where $E^{i}_{x_{i}}\in\mathcal{R}^m$ and $R^{i}_{x^{i}_{t}}\in\mathcal{R}^{d}$ are the $m$-dimensional feature embedding and the $d$-dimensional structure-aware embedding of the patch $I^{i}_{x^{i}_{t}}$, respectively.  $f^{i}_{t}$ and $u^{i}_{t}$ are the forget gate and the input gate, respectively. $C^{i}_{t}\in\mathcal{R}^{m+d}$ represents the hidden feature state. $S^{i}_{t}\in\mathcal{R}^3$ donates the hidden spatial state, and it can be regarded as the weighted sum of the previous landmark predictions. $\mathcal{W}^{i}_{\gamma}$ is a linear layer. $\mathcal{W}^{i}_{\gamma}C^{i}_{t}$ is predicted offset from $S^{i}_{t}$ to the ground truth landmark position. Therefore, $x^{i}_{t+1}\in\mathcal{R}^3$ is the sum of the predicted offset and the hidden spatial state. $\copyright$ is the concatenation operation. We present details of above notations in the following.

\begin{figure*}
  \centerline{\includegraphics[width=1\textwidth]{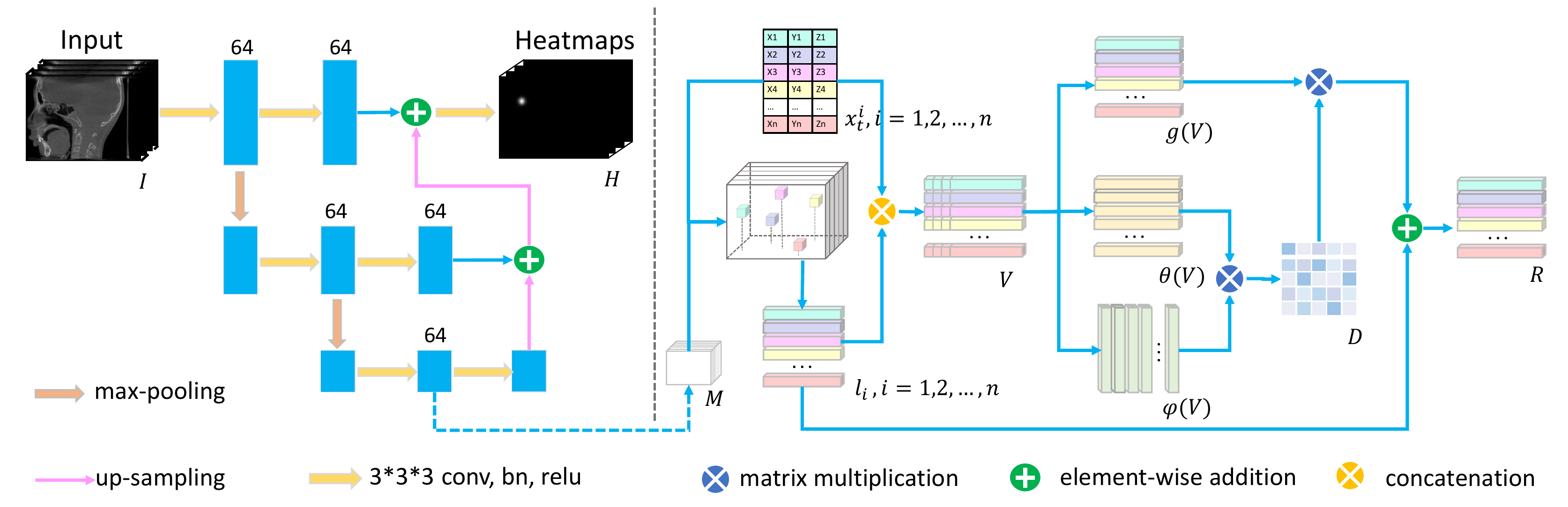}}
  \caption{The left-hand side is $\mathcal{F}{\alpha}$ which is used for coarse landmark detection, and the right-hand side is the graph attention module $\mathcal{F}{\gamma}$ which is used for calculating structure-aware embedding. The global feature vector $l_{i},i=1,2,...,n$ of $n$ landmarks are the feature vectors of the correspondence pixels in the coarse feature map $M$. The spatial  location $x^{i}_{t},i=1,2,...,n$ of $n$ landmarks are predicted from previous iteration. Moreover, the whole process is differentiable.}
  \label{fig:coarse_GAM}
  \vspace{-3ex}
\end{figure*}

\subsubsection{\textbf{Feature Embedding}} Visual Feature Embedding $E^{i}_{x^{i}_{t}}$ encodes the feature information of the 3D patch $I^{i}_{x^{i}_{t}}$, which is obtained by:
\begin{equation}\label{equ:embedding}
E^{i}_{x^{i}_{t}} = \mathcal{F}_{\beta}(I^{i}_{x^{i}_{t}}).
\end{equation}

Note that all patches with different resolutions are resized to be unified. $\mathcal{F}_{\beta}$ is a shared network for encoding all patches, containing four sequential sub-networks followed by an average pooling layer. Each sub-network contains a convolutional layer where the kernel size is 4$\times$4$\times$4; the stride is 2 and padding is 1, followed by the batch normalization and the ReLU activation (Figure \ref{fig:framework}).

\subsubsection{\textbf{Structure-aware Embedding}} Global structure constraint provides meaningful guidance to boost the performance. Based on this consideration, we propose a novel Structure-aware Embedding to implicitly encode the landmark's global structure information. Here the global structure represented by a graph $G = \{D, V\}$. The node set is $V = \{(v_{i} = x^{i}\copyright l_{i}) |i = 1, 2, ..., n\}$, where the spatial location $x^{i}\in\mathcal{R}^3$ of the $i$-th landmark is predicted in previous iteration. $l_{i}\in\mathcal{R}^{d}$ donates a d-dimensional feature vector of the corresponding pixel in the coarse feature map $M$ (Fig. \ref{fig:coarse_GAM}).  The spatial-visual feature $v_{i}$ is the concatenation of $x^{i}$ and $l_{i}$. The edge set $D=\{w(v_{i}, v_{j})|i=1,2,...,n;j=1,2,...,n\}$ contains learnable parameters. Inspired by \cite{vaswani2017attention,velivckovic2018graph}, we propose a graph attention module $\mathcal{F}_{\gamma}$ to propagate the spatial-visual feature to get the structure-aware embedding $R^{i}_{x^{i}_{t}}$ of each node:

\begin{equation}\label{equ:structure_embedding}
\begin{aligned}
&R^{i}_{x^{i}_{t}} = \mathcal{F}_{\gamma}(v_{i}) = l_{i} +  \frac{1}{U_{i}}\sum_{\forall j}w(v_{i}, v_{j})g(v_{j}), \\
&U_{i} = \sum_{\forall j}w(v_{i}, v_{j}),
\end{aligned}
\end{equation}
where $U_{i}$ is the normalization term, $g$ is a linear transformation function. The attention function $w$ is defined as a dot-product similarity between $v_{i}$ and $v_{j}$:
\begin{equation}\label{equ:structure_embedding}
w(v_{i}, v_{j}) = \theta(v_{i})^{T}\phi(v_{j}),
\end{equation}
where $\theta$ and $\phi$ donates the key and the query function, respectively.

\subsubsection{\textbf{Hidden Feature State and Hidden Spatial State}}

In conventional LSTM \cite{hochreiter1997long}, there is no spatial information for a hidden state, which results in inconsistent offset prediction. For example, If taking the conventional LSTM to predict offsets, we get the equation $x^{i}_{t+1} = \mathcal{W}^{i}_{\gamma}C^{i}_{t} + x^{i}_{t}$. However, the hidden feature state $C^{i}_{t}$ is obtained from previous hidden feature states, its spatial location is actually not the current patch central $x^{i}_{t}$. Therefore, the equation has an
inherent error. To perform consistent offset prediction, we introduce the spatial state $S^{i}_{t}$ to record the spatial information of $C^{i}_{t}$ (formula \ref{equ:spatialState}). 

The initial feature state $C^{i}_{1}$ is the concatenation of $E^{i}_{x^{i}_{1}}$ and $R^{i}_{x^{i}_{1}}$, and the initial spatial state $S^{i}_{1}$ is the coarse predicted landmarks $x^{i}_{1}$. To ensure the consistency, both $C^{i}_{t}$ and $S^{i}_{t}$ follow the same update rule, which is the weighted sum of the previous state and the current state (formula \ref{equ:featureState} and \ref{equ:spatialState}).
\subsubsection{\textbf{Self-attention Gates}} The accuracy of offset regression is sensitive to information the patch contained. For example, predicting offset using the patches cropped from the background is not reliable. Moreover, the accuracy is affected by the resolution of the patch size. For instance, a small patch has a limited perspective view. In contrast, a large patch has a wider perspective view but will lose the geometrical details if down-sampling. To exploit these local dependencies to meet individual landmark detection requirements, we introduce the self-attention mechanism to measure the relevance between the hidden feature state and the $i$-th landmark prediction. When updating the hidden feature state $C^{i}_{t}$ (formula \ref{equ:featureState}), the forget gate $f^{i}_{t}$ is used to determine how much information of the previous feature state $C^{i}_{t-1}$ should be maintained, and the input gate $u^{i}_{t}$ is used to determine how much information of a new input $E^{i}_{x^{i}_{t}}\copyright R^{i}_{x^{i}_{t}}$ should be accepted. By the iterative elimination and selection, the hidden feature state maintains the most relevant features. Thus, a more accurate prediction is aggregated accordingly. Both gates' values are normalized to be the range of $(0,1)$ and sum of 1, obtained by the following formula:

\begin{equation}\label{equ:attention}
\begin{aligned}
&f^{i}_{t} = \frac{\exp(\mathcal{A}^{i}(C^{i}_{t-1}))}{{K^{i}_{t}}},u^{i}_{t} = \frac{\exp(\mathcal{A}^{i}(E^{i}_{x^{t}_{i}}))}{K^{i}_{t}}, \\
&K^{i}_{t} = \exp(\mathcal{A}^{i}(C^{i}_{t-1}))+\exp(\mathcal{A}^{i}(E^{i}_{x^{t}_{i}})),
\end{aligned}
\end{equation}
where $K^{i}_{t}$ is a  normalization term. $\mathcal{A}^{i}(\cdot)$ is a self-attention network to measure the relevance between the input and the $i$-th landmark detection, which is formulated by the following:
\begin{equation}\label{equ:attention}
\mathcal{A}^{i}(X) =\sigma_{s}( \mathcal{P}^{i}_{\tau}\cdot \sigma_{t}(\mathcal{P}_{s}X)),
\end{equation}
where $\mathcal{P}_{s}$ is a shared linear layer, and $\mathcal{P}^{i}_{\tau}$ is a learnable network parameter allocated to the $i$-th landmark for calculating the relevance between the input $X$ and the $i$-th landmark detection. $\sigma_{s}$ is a Sigmoid activation function and $\sigma_{t}$ is a Tanh activation function.
\subsubsection{\textbf{Landmark Prediction}} 
By processing $T$ iterations, we get $T$ results $x^{i}_{t+1} (t=1,2,...,T)$ for the $i$-th landmark prediction. We take the results of the $T$-th iteration $x^{i}_{T+1}$ as the final predicted landmarks because it aggregates all previous predictions by attention gates. Essentially, $x^{i}_{T+1}$ can be regarded as the weighted sum of all previous $T$ predictions, where the learned weights measure the relevance between the hidden feature state and the $i$-th landmark prediction. $T$ is set to be $3$ to balance the inference time and accuracy.

\subsection{Training Details}

\subsubsection{\textbf{Multi-resolution Patch Cropping}}

We adopt a multi-resolution cropping strategy for data augmentation. Specifically, the size of 3D patches varies in different iterations of the fine-scale landmark detection stage, i.e., $I^{i}_{x^{i}_{1}}\geq I^{i}_{x^{i}_{2}}\geq I^{i}_{x^{i}_{3}}$ in terms of resolution. To this end, SA-LSTM aggregates the results under different perception fields for more precise prediction. Considering the predictions $x^{i}_{t}(i=1,2,...,n)(t=1,2,...,T)$ will converge to the ground truth $g_{i}(i=1,2,...,n)$, it will result in the lack of diversity of training samples if cropping patches centered on $x^{i}_{t}$. Therefore, 
during training SA-LSTM, we add Gaussian noise $\delta^{i}_{t} \sim N(0, \sigma_c) $ to the cropping centrals for $i$-th landmark in $t$-th iteration to simulate the actual inferring scenario. $\sigma_c$ is set to be $r_{t}/3$, $r_{t}$ is the radius of the cropping patches in $t$-th iteration, and 3 is determined based on the 3$\sigma$ principle of the Gaussian distribution, i.e., 99.73\% of patches contains the ground truth landmarks.

\subsubsection{\textbf{Loss function}}
In the coarse landmark detection stage, the loss function $\mathcal{L}_{h}$ for training the network $\mathcal{F}_{\alpha}$ is: 

\begin{equation}\label{equ:heatmapLossFunction}
\mathcal{L}_{h} = \sum_{i=1}^{n}\sum_{\rho\in \mathcal{H}_{i}}|\mathcal{H}_{i}(\rho) -  \exp{(\frac{|\rho - g_{i}|_2^2}{-2 * \sigma_h^2}})|,
\end{equation}
where $g_{i}$ is the ground truth of $i$-th landmark. $\mathcal{H}_{i}(\rho)$ is the value of the $i$-th heatmap $\mathcal{H}_{i}$ at the voxel $\rho\in\mathcal{R}^3$. The $\sigma_h$ is the standard deviation, and it is set as 2 empirically. In the fine-scale landmark detection stage, the loss function $\mathcal{L}_{h}$ for training SA-LSTM is as follows. 

\begin{equation}\label{equ:heatmapLossFunction}
\mathcal{L}_{c} = \sum_{i=1}^{n}\sum_{t=1}^{T}|x^{i}_{t} - g_{i}|.
\end{equation}

Therefore, the overall loss function is:
\begin{equation}\label{equ:overallLossFunction}
\mathcal{L}_{o} = \lambda\mathcal{L}_{h} + (1-\lambda)\mathcal{L}_{c},
\end{equation}
where $\lambda$ is the balance term, and set to be 0.5 empirically.

\begin{figure*}
  \centerline{\includegraphics[width=1\textwidth]{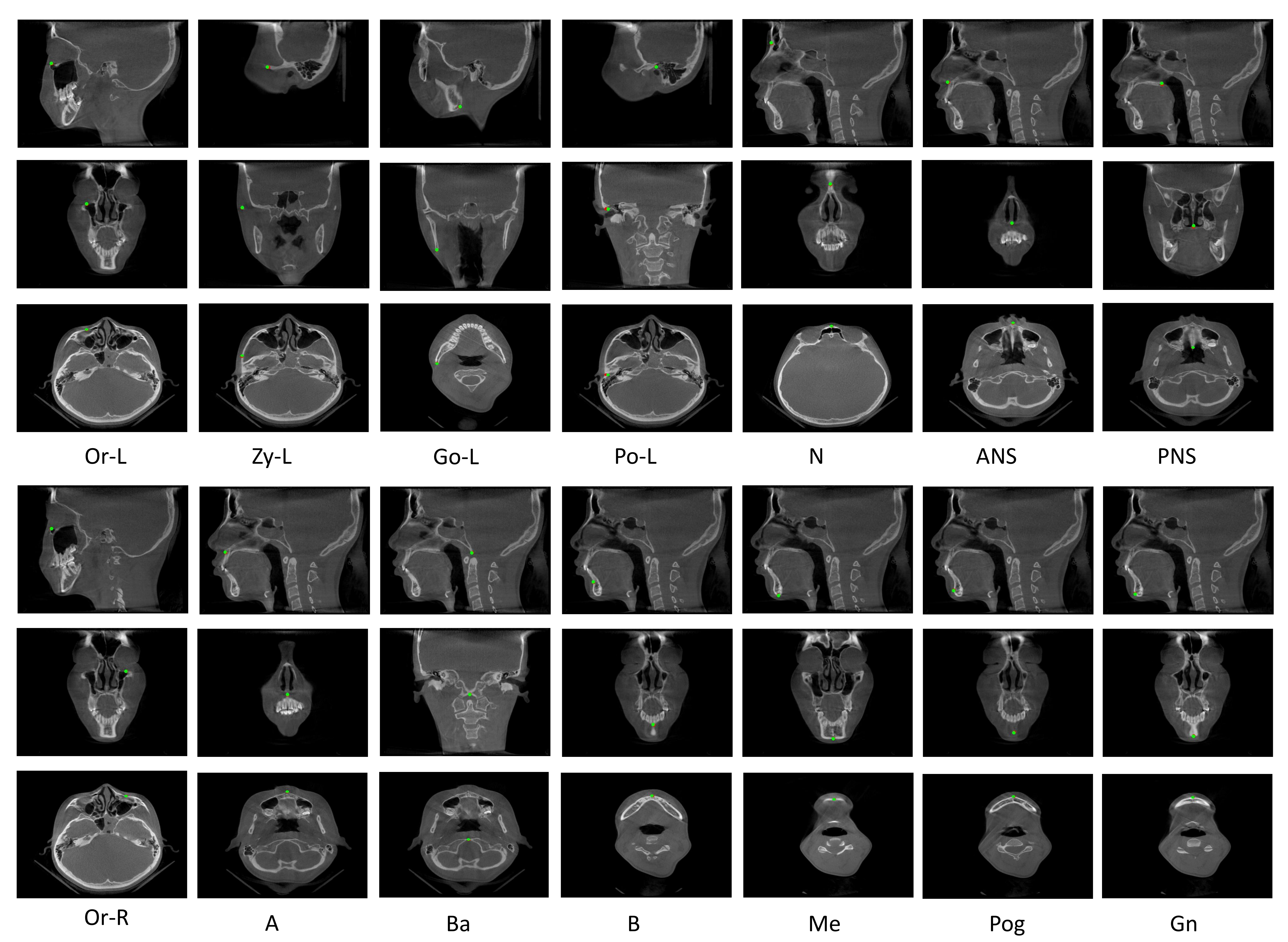}}
  \caption{Qualitatively evaluation of the detection results on our in-house dataset. Each landmark is presented as the X, Y, Z-axis projection views, respectively. Green points are the ground truth, and red points are predicted landmarks by our method.}
  \label{fig:visual}
\end{figure*}

\section{Experiments}
Our method is evaluated on two 3D volume datasets and outperforms current state-of-the-art methods. The advantages and the limitations of SA-LSTM are also discussed. In the end, we conduct ablation experiments to highlight the contribution of SA-LSTM and present its insights.

\subsection{Experiment Configuration}

Our method is evaluated on a in-house CBCT dataset (Fig. \ref{fig:data_sample}) and a public CT dataset \cite{raudaschl2017evaluation} by comparing with five state-of-the-art deep learning-based methods and a traditional regression-based method. The evaluation metrics are the mean radial error (MRE), standard deviation (SD), and the successful detection rate (SDR) in five target radius (2 mm, 2.5 mm, 3 mm, 4 mm, 8 mm) \cite{wang2016benchmark}. Besides, we also report the inference time for efficiency comparison.

\subsubsection{\textbf{In-house Skull Dataset}}
The dataset consists of 89 skull CBCT in DICOM format. For a CBCT, there are 576 2D slices with a resolution of 768 $\times$ 768. We transfer all CBCT data to 3D volumes with the size of 768 $\times$ 768 $\times$ 576. The isotropic voxel spacing is 0.3 mm. Each data sample has 17 manually annotated landmarks on the skull (Fig. \ref{fig:data_sample}). The dataset is distributed into three folds with 29, 30 and 30 samples to perform three-fold cross-validation. We report the average accuracy of three folds. 

\subsubsection{\textbf{PDDCA Dataset}}
PDDCA dataset is a Public Domain Database for Computational Anatomy. It contains 48 patient CT images from the Radiation Therapy Oncology Group (RTOG) 0522 study, with manual segmentation of multiple organs and manual identification of bony landmarks. It was been used for MICCAI Head-Neck Challenge 2015 \cite{raudaschl2017evaluation}, and 33 CT images are labeled with five bony landmarks, i.e., Chin (chin), Right condyloid process (mand\_r), Left condyloid process (mand\_l), Odontoid process (odont\_proc), Occipital bone (occ\_bone). The dataset is equally distributed into three folds to perform three-fold cross-validation, and the average accuracy of three folds is reported. Image files are stored in compressed NRRD format, and the spacing range for $x$, $y$ and $z$ axis are 0.8 mm$\sim$1.2 mm, 0.8 mm$\sim$1.2 mm and 2.5 mm$\sim$3.3 mm, respectively. To standardize the input data, we resize the image to ensure the isotropic voxel spacing is 1 mm. Therefore, the average resolution is 518 $\times$ 518 $\times$ 384 after standardization.

\subsubsection{\textbf{Implementation Details}}
The model configuration is the same for all experiments. The architecture of $\mathcal{F}{\alpha}$ for coarse landmark detection is shown in Fig. \ref{fig:coarse_GAM}, and the filter number $d$ is set to be 64 for all convolutional layers. For patch feature encoder $\mathcal{F}_{\beta}$ with four convolutional layers, the filter numbers are 64, 128, 256 and 512, respectively. The input and output's dimensions number for the offset predictor $\mathcal{W}_{\gamma}^{i}$ are 576 and 3, respectively. In the self-attention network $\mathcal{A}^{i}(\cdot)$, the dimension of the input, inter-medial, and output are 576, 256 and 1, respectively.

In the first stage, the size of the down-sampled volume for the In-house dataset and PDDCA dataset are 96 $\times$ 96 $\times$ 72 and 64 $\times$ 64 $\times$ 48, respectively, is approximately 1/8 of their original size. There is a trade-off between efficiency and accuracy, i.e., the smaller patch cost little computational resources but provides less accurate coarse landmarks. The patch size are respectively 64 $\times$ 64 $\times$ 64, 32 $\times$ 32 $\times$ 32, 16 $\times$ 16 $\times$ 16 for three iterations in fine-scale prediction. All patches with different resolutions are unified resized to 32 $\times$ 32 $\times$ 32 as the input of feature encoder $\mathcal{F}_{\beta}$. The dimension of the feature embedding $m$ is 512.

Networks are trained 1000 epochs by Adam optimizer \cite{kingma2015adam} with the default configuration in PyTorch platform. The training time is approximately 20 hours and 3 hours for the in-house dataset and PDDCA dataset, respectively, on a GTX 1080 Ti GPU.

\subsubsection{\textbf{Baselines}}
Due to some state-of-the-art 3D anatomical landmark detection methods rely on the segmented bone \cite{lee2019automatic,zhang2015automatic,torosdagli2018deep,lian2020multi}, our method is not directly compared with these methods. To verify the superiority of our method, we re-implement five deep learning-based methods and a traditional regression-based method for extensive comparison, i.e. 3D U-net \cite{cciccek20163d}, SCN \cite{Payer2019Integrating}, DRM \cite{zhong2019attention}, LA-GCN \cite{lang2020automatic}, ADC \cite{chandran2020attention} and CoRe \cite{gao2015collaborative}. 3D U-net is a volumetric segmentation network that learns sparse annotations. We modified the last layer for predicting landmark's heatmaps and calculating landmark positions by integral operation (formula \ref{equ:integralFunction}). SCN is a state-of-the-art method for 3D landmark detection in Hand CBCT and Spine CBCT. Since the original volume is too big to be processed by these two networks, the same down-sampled volume to ours is used as network input. For a fair comparison, we also show the performance against three coarse-to-fine methods. DRM is a state-of-the-art method for 2D cephalometric landmark detection. LA-GCN is a state-of-the-art method for 3D Craniomaxillofacial landmark detection in CBCT. We do not use spiral CT images to pre-train the model for a fair comparison. ADC is the state-of-the-art method for high-resolution facial landmark detection. CoRe is a traditional regression-based anatomical landmark detection method on 3D CT/CBCT. These two-stage methods share the same hyper-parameter as our method, e.g., patch size, down-sampled volume size.

\subsection{Results and Discussions}
We first quantitatively compare our method with other state-of-the-art methods and then discuss its superiority in terms of accuracy, robustness, and efficiency.
\subsubsection{\textbf{In-house Skull Dataset}}
As shown in Table \ref{tab:inin-house}. Our method outperforms the state-of-the-art methods. Especially in the precision of 2 mm $\sim$ 3 mm, the improvements of SDR are 7\%, 2\% and 1\%, respectively. The average error of three-fold cross-validation is 1.64 mm, and 74.28\% of landmarks are within the 2 mm clinically acceptable errors. The individual landmark accuracy is shown in Table \ref{tab:landmarks}. 100\% of landmarks are detected within 8 mm errors. 94.38\% of the landmark Ba and 95.50\% of the landmarks ANS are detected within 2 mm errors. The inference time of our method is 0.46 seconds for 17 landmark detection on the original-sized CBCT, significantly faster than the other two-stage methods (ADC, DRM, LA-GCN, CoRe). The quality evaluation of the detection results is presented in Figure \ref{fig:visual}, showing the accuracy of the predicted landmarks is comparable to human annotations.

\begin{table*}[h]
	\centering
	\caption{Accuracy of our method for individual landmarks. We present the accuracy of 17 landmarks as well as the average accuracy of three-folds cross-validation measured by MRE, SD and SDR.}\label{tab:landmarks}
	\scalebox{1}{
	\begin{tabular}{l|c c c c c c|l|c c c c c c}
		\hline
		
        \hline
		\multirow{2}*{\tabincell{c}{Landmark}} & \multirow{2}*{\tabincell{c}{MRE(SD)}} & \multicolumn{5}{c|}{SDR} & \multirow{2}*{\tabincell{c}{Landmark}} & \multirow{2}*{\tabincell{c}{MRE(SD)}} & \multicolumn{5}{c}{SDR}\\
        \cline{3-7}
        \cline{10-14}
		~ & ~& 2mm & 2.5mm & 3mm & 4mm & 8mm & ~ & ~& 2mm & 2.5mm & 3mm & 4mm & 8mm\\

        \hline

		Or-R& 1.59(0.95) & 70.78 & 82.02 & 89.88 & 98.87 & 100 &
		Or-L& 1.40(0.95) & 82.02 & 92.13 & 93.25 & 96.62 & 100 \\
		Zy-R& 2.21(1.64) & 51.68 & 64.04 & 73.03 & 85.39 & 100 &
		Zy-L& 1.97(1.43) & 65.16 & 68.53 & 75.28 & 88.76 & 100 \\
		Go-R& 2.20(1.63) & 51.68 & 66.29 & 73.03 & 89.88 & 100 &
		Go-L& 2.28(1.54) & 55.05 & 66.29 & 73.03 & 85.39 & 100 \\
		Po-R& 1.69(1.04) & 70.78 & 78.65 & 88.76 & 94.38 & 100 & 
		Po-L& 1.61(1.04) & 76.40 & 85.39 & 91.01 & 94.38 & 100 \\
		N & 1.69(1.34) & 73.03 & 84.26 & 88.76 & 92.13 & 100  & 
		ANS & \textbf{0.89(0.52)} & 95.50 & 98.87 & 100 & 100 & 100 \\
		PNS & 1.26(0.90) & 88.76 & 93.25 & 95.50 & 96.62 & 100  &
		A& 1.38(0.86) & 84.26 & 92.13 & 94.38 & 97.75 & 100 \\
		Ba& \textbf{0.98(0.64)} & 94.38 & 97.75 & 97.75 & 98.87 & 100  & 
		B& 2.17(1.64) & 61.79 & 67.41 & 76.40 & 87.64 & 100 \\
		Me & 1.32(0.86) & 87.64 & 88.76 & 94.38 & 98.87 & 100  & 
		Pog & 1.67(1.16) & 74.15 & 84.26 & 91.01 & 93.25 & 100 \\
		Gn  & 1.54(1.05) & 79.77 & 86.51 & 87.64 & 95.50 & 100  &
		Average &1.64(1.13) & 74.28 & 82.15 & 87.24 & 93.78 & 100\\

	\end{tabular}}
\end{table*}

\subsubsection{\textbf{PDDCA Dataset}}
PDDCA dataset has a much lower resolution than our in-house dataset (0.8 mm$\sim$3.5 mm spacing VS 0.3 mm spacing) and has fewer samples (33 samples VS 89 samples). However, our method is robust to such a dataset. As shown in Table \ref{tab:inin-house}, our method significantly outperforms other deep learning-based methods. The improvement of SDRs is increased by 20\%$\sim$25\% in the precision of 2 mm $\sim$ 4 mm. The average error of three folds is 2.37 mm, and 89.99\% of landmarks are within the 4 mm errors. However, our method is comparable with CoRe because the dataset may not be challenging enough and bony landmarks on CT images have a distinctive appearance, which handcraft features can well model. Note that ADC achieves deficient performance on this dataset, so the result is not reported. The inference time is 0.27 seconds for five landmark detection on the original-sized CBCT.

\subsubsection{\textbf{Discussions}}
\paragraph{Compared with 3D U-net and SCN}
3D U-net is a baseline method that aggregates the feature pyramid for medical image segmentation. SCN is an improved version of the U-net that integrates spatial configuration for landmark localization. Even though these two methods are slightly faster than ours, they are unsuitable for high-resolution input data, thus are failing to capture the landmark's geometric details. Therefore, our method significantly more accurate than these methods. 

\begin{table*}[h]
	\centering
	\caption{Comparison with state-of-the-art methods on in-house and PDDCA dataset and  by evaluation metrics of Time, MRE, SD, and SDR in five target radius.}\label{tab:inin-house}
	\scalebox{1}{
	\begin{tabular}{l|l|c c c c c c c}

        \hline
        \multirow{2}*{\tabincell{c}{Dataset}} &
		\multirow{2}*{\tabincell{c}{Model}} &
		\multirow{2}*{\tabincell{c}{Time}} &
		\multirow{2}*{\tabincell{c}{MRE(SD)}} & \multicolumn{5}{c}{SDR}\\
        \cline{5-9}
		~&~ & ~&~& 2mm & 2.5mm & 3mm & 4mm & 8mm\\

        \hline
		\multirow{7}*{\tabincell{c}{In-house}}&3D U-Net\cite{zheng20153d}&0.34s& 2.47(1.47) & 46.02 & 62.42 & 74.66 & 87.99 & 99.41\\
		~&SCN \cite{Payer2019Integrating}&0.48s& 2.30(1.24) & 49.98 & 65.81 & 76.32 & 89.18 & 99.54\\
		~&ADC \cite{chandran2020attention}&10.31s& 3.67(1.76) & 27.45 & 36.27 & 48.43 & 62.74 & 94.70\\
		~&DRM \cite{zhong2019attention}&21.64s& 1.98(1.29) & 61.17 & 71.76 & 80.19 & 91.96 & 99.61\\
		~&LA-GCN \cite{lang2020automatic}&184.35s& 1.85(1.21) & 65.34 & 74.83 & 83.24 & 92.13 & 99.61\\
		~&CoRe \cite{gao2015collaborative} & 23.31s & 1.75(1.13) & 67.64 & 80.19 & 86.27 & \textbf{93.92} & 99.80 \\
		~&\textbf{SA-LSTM}&\textbf{0.46s}&\textbf{1.64(1.13)} & \textbf{74.28} & \textbf{82.15} & \textbf{87.24} & 93.78 & \textbf{100} \\
		
		\hline
		\multirow{6}*{\tabincell{c}{PDDCA}}&3D U-Net\cite{zheng20153d}&0.16s&7.69(5.24) & 2.03 & 3.25 & 5.05 & 16.28 & 67.90\\
		~&SCN \cite{Payer2019Integrating}&0.25s& 7.44(4.26) & 2.65 & 6.74 & 10.98 & 21.36 & 69.30\\
		~&DRM \cite{zhong2019attention}&12.32s& 6.39(3.37) & 7.27 & 12.72 & 16.36 & 29.09 & 74.54\\
		~&LA-GCN \cite{lang2020automatic}&93.21s& 3.23(2.52) & 35.68 & 46.76 & 58.19 & 69.48 & 94.74\\
		~&CoRe \cite{gao2015collaborative} & 17.47s & \textbf{2.16(1.02)} & 53.49 & \textbf{73.86} & \textbf{84.57} & 88.82 & 94.95 \\
		~&\textbf{SA-LSTM}&\textbf{0.27s}&2.37(1.60) & \textbf{56.36} & 71.60 & 80 & \textbf{89.99} & \textbf{95.91} \\
	\end{tabular}}
\end{table*}

\paragraph{Compared with DRM}
DRM is a two-stage heatmap-based method for 2D cephalometric landmark detection. Similar to our method, they first predict the coarse landmarks on a down-sampled image and then ensemble the predictions of the fine-scale landmarks based on the cropped patches. However, they have some major weaknesses when compared with our method. Firstly, they perform heatmap regression for fine-scale landmark detection, which results in an unreliable prediction when the cropped patch does not contain the landmark. Our method predicts the offset between the patch central and the ground truth landmark, which means that there is no need for the patch to contain the landmark. Experiments show that our method is robust to the coarse landmark accuracy. Secondly, in their method, the ensemble of multiple regressions is computationally expensive in processing 3D volume data (taking half of the minutes for one iteration and always exceeding 11G GPU memory of commonly used GTX 1080 TI). While in our method, by using the iterative search strategy, the SA-LSTM is very efficient as it only needs three small patches to detect each landmark (detecting all landmarks in 0.5s with 0.5G GPU memory used). Therefore, our method is more efficient to handle 3D data. Thirdly, due to each landmark being independently predicted in the second stage of their method, some landmarks may have a great deviation from the ground truth, which results in unacceptable diagnosis in cephalometric analysis. We solve this problem by introducing a novel Structure-aware Embedding to implicitly encode the global structure information. Lastly, although averaging the multiple regressions seems a reasonable strategy in their method, it is sensitive to some bad cases. For example, some patches cropped from the background, or some patches do not contain landmarks. Averaging these low-quality predictions instead decrease the final accuracy. In our method, after the iterative elimination and selection by attention gates, SA-LSTM maintains the most relevant features. Thus, a more accurate prediction is performed.

\paragraph{Compared with ADC}
ADC directly learns the image-to-landmarks mapping in a coarse-to-fine manner. Although it achieves state-of-the-art performance on high-resolution facial landmark datasets, it does not perform well in the 3D CBCT dataset. Because it is difficult to directly learn the image-to-landmarks mapping with only a few training data in 3D form (60 samples). 

\paragraph{Compared with LA-GCN}
LA-GCN is a two-stage approach to localize landmarks via 3D heatmap regression. They utilize mask R-CNN to detect the landmarks by multi-task learning under different resolutions. For a fair comparison, we share the same network architecture in coarse landmark detection. As shown in Table \ref{tab:inin-house}, their method is computationally expensive (it takes minutes to infer a CBCT volume), while the inference time of our SA-LSTM is within 0.5 seconds, which is much more efficient. Besides, benefit from the structure regulation by structure-aware embedding and the feature aggregation by attention gates, our method achieves more accurate and robust results.

\paragraph{Compared with CoRe} CoRe is a traditional regression-based method that models landmark's appearance by Haar-like features. It trains a regression forest to predict the offsets from the positions sampled by spherical sampling strategy, and aggregate the final result by collaborative regression voting, i.e., using easy-to-detect landmarks to guide the detection of difficult-to-detect landmarks. Experiments show that it is comparable with our method in the PDDCA dataset. It is mainly because that the dataset may not be challenging enough, and handcraft features can well handle the bony landmarks' appearance on CT images. Our method achieve better performance in the in-house dataset (Table \ref{tab:inin-house}). Since the samples of the in-house dataset are more diverse and complex than the PDDCA dataset (89 CBCT with spacing 0.3 mm VS 33 CT with spacing 0.8mm $\sim$ 3.5mm), our deep learning-based framework is more suitable for handling a more complex and fine-grained landmark detection task.

\begin{table*}[h]
	\centering
	\caption{Ablation experiments on our in-house dataset by evaluation metrics of MRE, SD, and SDR in eight target radius.}\label{tab:ablation}
	\scalebox{1}{
	\begin{tabular}{l|c c c c c c c c c}
		\hline
		
        \hline
        \multirow{2}*{\tabincell{c}{Config}} & \multirow{2}*{\tabincell{c}{MRE(SD)}} & \multicolumn{8}{c}{SDR}\\
        \cline{3-10}
        ~ & ~& 1mm&2mm&3mm&4mm&5mm&6mm&7mm&8mm\\
        \hline
        Base&2.25(1.15)&12.74&50.71&79.84&91.13&95.25&97.16&98.89&99.48 \\
        Base$\dagger$\_I-3&1.78(1.19)&31.18&71.54&87.75&93.8&95.15&97.86&99.21&99.6 \\
        Base+AGN\_I-3&1.73(1.16)&31.94&72.53&88.53&93.48&95.88&97.95&99.33&99.48 \\
        Base+GAM\_I-3&1.62(1.01)&33.52&73.33&90.19&94.49&96.45&99.41&99.8&99.8 \\
        Base+SAG\_I-3&1.66(1.15)&33.92&74.31&89.21&94.31&96.86&98.03&99.41&99.41 \\
        \hline

        \hline
        Base+GAM+SAG(w/o h) & 1.71(1.16) & 33.70 & 70.19 & 88.03 & 93.52 & 97.05 & 98.43 & 99.41 & 99.60 \\
        Base+GAM\_I-1&1.62\textbf{(0.97)}&31.76&73.92&89.6&94.7&97.25&99.21&99.6&99.8 \\
        Base+GAM+SAG\_I-2&1.61(1.02)&32.74&74.31&90.19&94.7&97.25&99.21&99.6&100 \\
        \textbf{Base+GAM+SAG\_I-3}&\textbf{1.60}(1.03)&\textbf{34.31}&\textbf{75.29}&\textbf{90.19}&\textbf{95.9}&\textbf{98.25}&\textbf{99.6}&\textbf{100}&\textbf{100} \\
		
	\end{tabular}}
	\vspace{-2ex}
\end{table*}

\paragraph{Compared with other related methods}
AGN \cite{schlemper2019attention} proposed another attention gates for medical image classification and segmentation. However, some significant differences make our method more accurate (Table \ref{tab:ablation} in Ablation experiments). In motivation, their attention weights are calculated on feature maps that suppressing feature responses in irrelevant background regions. In our method, the attention weights are calculated among multiple predictions, progressively maintaining the relevant features and eliminating irrelevant features for the final result aggregation. For the implementation details, they utilize contextual semantic features to calculate the attention weight shared for all classes. While in our method, we allocate the learnable network parameter $\mathcal{P}^{i}_{\tau}$ for each type of landmark to learn fine-grained attention weight. As shown in Table \ref{tab:ablation} (Base+AGN\_I-3 VS Base+SAG\_I-3), our attention gates achieve better performance. 

Unlike CNN to encodes local features with the convolutional operation, Vision Transformer \cite{carion2020end,li2021medical,khan2021transformers} models the pixel-wised dependency via self-attention, achieving promising results in various computer vision tasks. However, modelling pixel-wise dependency is very computationally expensive. In our experiment, the maximal input size to the Transformer with eight heads six blocks is 16$\times$16$\times$16 on an 11G GTX 1080TI GPU. Therefore, the Transformer can hardly capture both long-range and short-range pixel-wised dependence across a large-scaled CBCT volume. Our method adopts self-attention to model the long-range and short-range dependency among cropped patches by the graph attention module and self-attention gates, respectively, overcoming the Transformer's limitation. However, the parallel computing exiting in Transformer seems more efficient than the recursive prediction, inspiring us to improve our method's efficiency in future work.

\begin{table*}[h]
	\centering
	\caption{The performance under the different down-sampled sizes of the coarse stage on the in-house dataset by evaluation metrics of Time, MRE, SD, and SDR in five target radius.}\label{tab:resulotion}
	\scalebox{1}{
	\begin{tabular}{l|l|c c c c c c c}
		\hline
		
        \hline
		\multirow{2}*{\tabincell{c}{Down-sampling Size}} &
		\multirow{2}*{\tabincell{c}{Model}} &
		\multirow{2}*{\tabincell{c}{Time}} &
		\multirow{2}*{\tabincell{c}{MRE(SD)}} & \multicolumn{5}{c}{SDR}\\
        \cline{5-9}
		~ &~ & ~&~& 2mm & 2.5mm & 3mm & 4mm & 8mm\\

        \hline
		\multirow{2}*{\tabincell{c}{48*48*32}} &
		Coarse&0.06s &3.48(1.68) & 21.56& 35.09& 48.03& 67.84& 96.47\\
		~ &SA-LSTM &0.21s& \textbf{1.55(0.96)} & \textbf{75.68} &84.70 & 91.76 & 96.05 & \textbf{100}\\

        \hline
		\multirow{2}*{\tabincell{c}{64*64*48}} & Coarse&0.12s &2.71(1.35) & 36.67 & 52.35 & 68.23 & 83.52 & 99.02\\
		~ &SA-LSTM &0.26s& 1.58(0.97) & 74.70 & 85.09 & 90.19 & 96.27 & \textbf{100}\\
		\hline
		\multirow{2}*{\tabincell{c}{88*88*64}} & Coarse&0.23s& 2.42(1.24) & 44.70 & 61.17 & 74.51 & 89.41 & 99.80\\
		~ &SA-LSTM &0.37s& 1.58(0.97) & 74.31 & 84.90 & \textbf{91.96} & \textbf{96.47} & \textbf{100}\\
		\hline
		\multirow{2}*{\tabincell{c}{96*96*72}} & Coarse&0.31s& 2.25(1.15) & 50.71 & 68.84 & 79.84 & 91.13 & 99.48\\
		~ & SA-LSTM &0.46s& 1.60(1.03) & 75.29 & \textbf{85.21}& 90.19 & 95.9 & \textbf{100} \\
		\hline
		\multirow{2}*{\tabincell{c}{128*128*96}} & Coarse&0.62s& 2.29(1.19) & 50.19 & 67.64 & 78.23 & 89.21 & 99.41\\
		~ & SA-LSTM &0.76s& 1.61(1.02) & 74.72 & 84.17 & 90.80 & 95.70 & \textbf{100}\\
		
	\end{tabular}}
	\vspace{-2ex}
\end{table*}

\paragraph{Limitation and Future work}
Our method is accurate, efficient and robust to perform landmark detection on 3D CBCT data. However, our method is unsuitable for processing incomplete input images, e.g., some landmarks are missing in input images. We will solve it in future work. Besides, Since bony landmarks on CT images have a distinctive appearance, our method's potential may not be fully shown on such datasets. Therefore, we plan to annotate the tooth landmarks in our dataset and develop a more powerful tooth landmark detection method based on SA-LSTM.

\subsection{Ablation Experiments}
To highlight the effectiveness of different modules in our framework, we conduct the following experiments. The baseline is the first stage of our method for the coarse landmark detection(Base). To verify the effectiveness of SA-LSTM in the second stage, we replace it with $T$ times naive iterative predictions (Base$\dagger$\_I-3). To verify the function of attention gates, we conduct $T$ times recursive offset predictions with self-attention gates (Base+SAG\_I-3). We also compared with other attention gates \cite{schlemper2019attention}. Specifically, we apply their attention gates in our feature embedding network $\mathcal{F}_{\beta}$ for $T$ times fine-scale landmark detection (Base+AGN\_I-3). We show the performance with structure-aware embedding (Base+GAM\_I-3) and our full method (Base+GAM+SAG\_I-3). Besides, the effectiveness of the number of iterations is presented(I-1, I-2, I-3). We conduct a experiment (Base+GAM+SAG(w/o h)) that remove the hidden spatial state (replaced $S^{i}_{t}$ with $x^{i}_{t}$ in formula \ref{equ:prediction}). Furthermore, we report the performance under different coarse landmark accuracy to show the robustness of SA-LSTM (Table \ref{tab:resulotion}). All experiments are conducted on our in-house CBCT dataset (59 samples for training and 30 samples for validation). We run five times per experiment and report the average of the highest performance of five times for each experiment.

\subsubsection{\textbf{Effect of SA-LSTM}}
As shown in Table \ref{tab:ablation}, significant improvements can be seen by applying our SA-LSTM for fine-scale landmark detection (Base VS Base+GAM+SAG\_I-3). Especially in the precision of 1 mm$\sim$3 mm, the improvements of SDR are about 32\%, 25\% and 11\%, respectively. It shows that SA-LSTM fully uses landmark geometrical details to improve the fine-scale detection performance. To further verify the effectiveness of its design, we replace SA-LSTM with $T$ times naive iterative offset predictions in the second stage. With the effort of the structure-aware embedding and the attention gates, the improvements of SDR are 2\%$\sim$4\% in the precision of 1 mm$\sim$6 mm (Base†\_I-3 VS Base+GAM+SAG\_I-3). More importantly, SA-LSTM eliminates the influence of some dire predictions and lifts the SDR from 99.41\% to 100\% within 7 mm precision. Lastly, observe from the experiment (Base+GAM+SAG(w/o h)) that removes the hidden spatial state (replaced $S^{i}_{t}$ with $x^{i}_{t}$ in formula \ref{equ:prediction}). As shown in Table \ref{tab:ablation} (Base+GAM+SAG(w/o h) VS Base+GAM+SAG\_I-3), the performance is dropped.

\subsubsection{\textbf{Effect of Self-attention Gates}}
Observing the experiment of Base†\_I-3 and Base+SAG\_I-3, Self-attention gates contribute 1\%$\sim$3\% improvement of SDR in the precision of 1 mm$\sim$5 mm. Intuitively, the confidences of each landmark prediction are measured by the self-attention mechanism, and by the iterative elimination and selection, the hidden feature state maintains the most relevant features. Thus, a better prediction is performed (Base+GAM+SAG\_I-1, I-2, I-3). Moreover, compared with other attention gates (Base+SAG\_I-3 VS Base+SGN\_I-3), our method is more efficient to alleviate the adverse effects caused by the irrelevant features.

\subsubsection{\textbf{Effect of Structure-aware Embedding}}
It can be seen from the experiment of Base†\_I-3 and Base+GAM\_I-3 that by aggregating structure-aware embedding for offset prediction, the performance under all evaluation metrics is improved. Notably, the SDRs are increased by 1\%$\sim$2\% in the precision of 1 mm$\sim$4 mm. Moreover, compared to our full method, the accuracy will reduce if without GAM (Base+GAM\_I-3 VS Base+GAM+SAG\_I-3), and some of the predicted landmarks have a significant deviation from the ground truth (0.59\% of prediction errors are larger than 8 mm).

\subsubsection{\textbf{Effect of Coarse Landmark Accuracy}}
As shown in Table \ref{tab:resulotion}, the smaller down-sampled size provides less accurate coarse landmarks with less inference time. Experiments show that SA-LSTM achieves comparable performance under different coarse landmark accuracy. Besides, to see the performance when the patches do not contain ground truth, we manually add a large bias (9$\sim$12mm) to the coarse landmarks for cropping patches. We observe that the accuracy is comparable to the counterpart which not do so (1.68 mm MRE VS 1.64 mm MRE). The above experiments show the robustness of SA-LSTM to the low accuracy of the coarse landmark.

\section{Conclusion}
We propose a novel structure-aware LSTM framework for efficient and accurate landmark detection on 3D CBCT volume. Our method is efficient as it detects the coarse landmarks on the down-sampled CBCT volume and progressively refines landmarks by attentive offset regression using the multi-resolution cropping patches. Besides, it boosts accuracy by capturing the global-local dependence among patches via self-attention. Specifically, it implicitly encodes the global structure information to make the predicted landmarks more reasonable. It recursively filters irrelevant local features and maintains the high-confident local prediction for aggregating the final result. Extensive experiments show that our method outperforms state-of-the-art methods in terms of accuracy and efficiency.

\bibliographystyle{IEEEtran}
\bibliography{egbib}

\end{document}